\documentclass[10pt,twocolumn,letterpaper]{article}
\usepackage{multirow}
\usepackage{iccv} 
\usepackage{bm}
\definecolor{cvprblue}{rgb}{0.21,0.49,0.74}
\usepackage[pagebackref,breaklinks,colorlinks,allcolors=cvprblue]{hyperref}

\def\paperID{55}
\def\confName{3DV}
\def\confYear{2026}

\title{Learning High-Fidelity Cloth Animation via Skinning-Free Image Transfer}

\author{
Rong Wang \quad Wei Mao \quad Changsheng Lu \quad Hongdong Li\\
The Australian National University\\
{\tt\small \{rong.wang, changsheng.lu,hongdong.li\}@anu.edu.au \quad wei.mao.research@gmail.com}
}

\begin{document}
\maketitle

\begin{abstract}
We present a novel method for generating 3D garment deformations from underlying body poses, which is key to a wide range of applications, including virtual try-on and extended reality. To simplify the cloth dynamics, existing methods mostly rely on linear blend skinning to obtain low-frequency posed garment shape and only regress high-frequency wrinkles. However, due to the lack of explicit skinning supervision, such skinning-based approach often produces misaligned shapes when posing the garment, consequently corrupts the high-frequency signals and fails to recover high-fidelity wrinkles. To tackle this issue, we propose a \textbf{skinning-free} approach by independently estimating posed (i) \textbf{vertex position} for low-frequency posed garment shape, and (ii) \textbf{vertex normal} for high-frequency local wrinkle details. In this way, each frequency modality can be effectively decoupled and directly supervised by the geometry of the deformed garment. To further improve the visual quality of deformation, we propose to encode both vertex attributes as \textbf{rendered} texture images, so that 3D garment deformation can be equivalently achieved via 2D image transfer. This enables us to leverage powerful pretrained image models to recover fine-grained visual details in wrinkles, while maintaining superior scalability for garments of diverse topologies without relying on manual UV partition. Finally, we propose a multimodal fusion to incorporate constraints from both frequency modalities and robustly recover deformed 3D garments from transferred images. Extensive experiments show that our method significantly improves animation quality on various garment types and recovers finer wrinkles than state-of-the-art methods.
\end{abstract}

\begin{figure}[htp!]
\centering  {\includegraphics[width=0.5\textwidth]{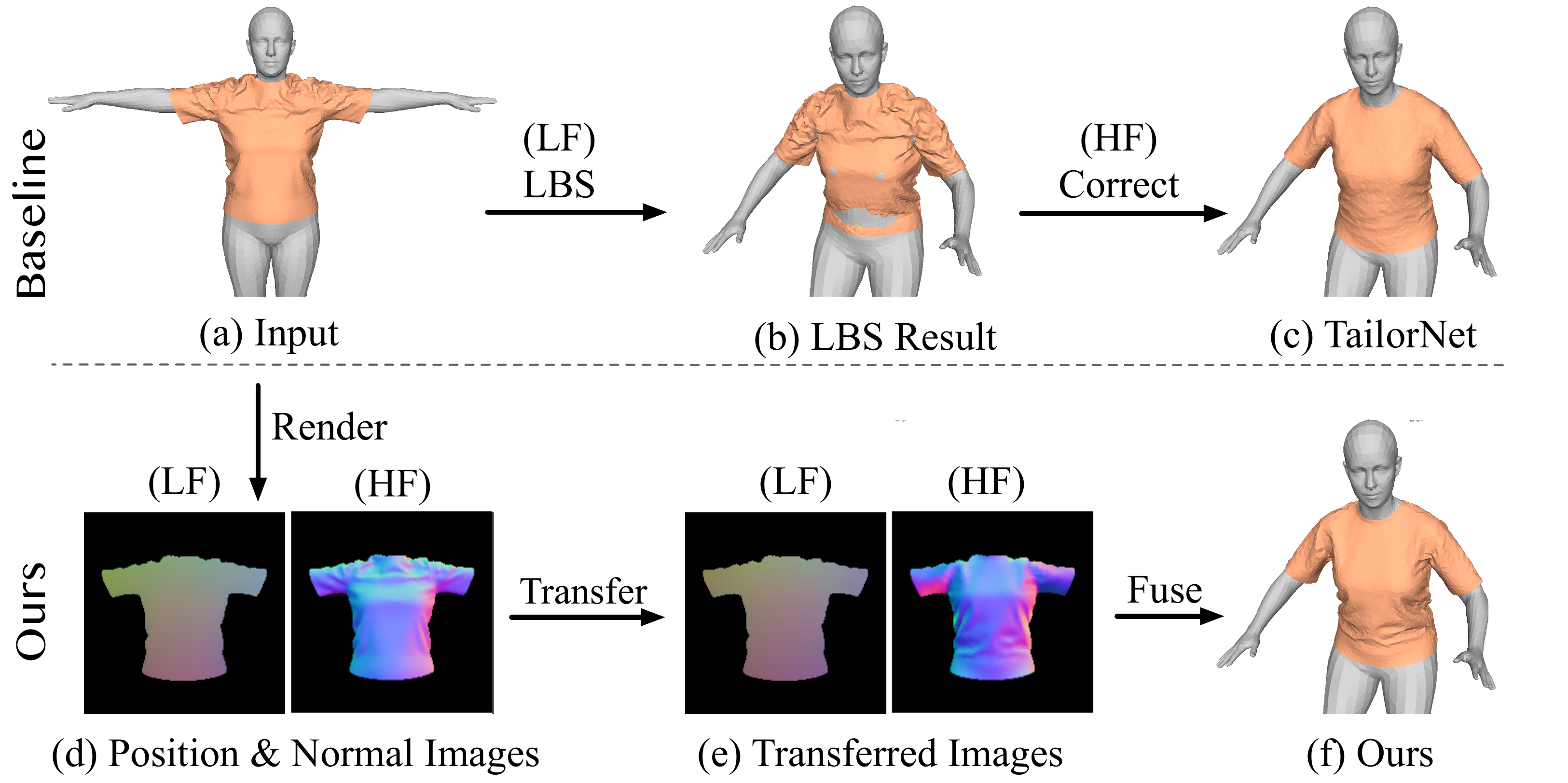}}
    \caption{\textbf{Illustration of our method.} Given input garment and body meshes (a), previous work \citep{patel2020tailornet} relies on LBS to generate low-frequency (LF) posed garment shape. However, inaccurate skinning in LBS can produce artifacts and misaligned garment position (b), which corrupts high-frequency (HF) signals and hinders the wrinkle regression (c). In contrast, we decompose frequency modalities using two geometric attributes: vertex positions and normals, which are rendered as 2D texture images (d) and then transferred on pixel intensities (e) to represent garment deformation. After fusing from both modalities, we generate deformed garment with more accurate wrinkles (f).}
    \label{fig:open}
    \vspace{-10pt}
\end{figure}

\section{Introduction} Generating high-quality 3D garment deformation given posed human bodies facilitates a wide range of applications, such as digital humans, virtual try-on, and extended reality. Traditional works \citep{provot1995deformation, baraff2023large} mostly rely on simulation to generate plausible results. However, such physics-based methods are time-consuming and require manual fine-tuning simulator settings for each garment, which is laborious and does not scale to diverse garment types \citep{luible2008simulation, zhang2024estimating}.

Recently, learning-based methods \citep{santesteban2019learning, patel2020tailornet, zhang2022motion, santesteban2022snug, pan2022predicting, ma2022neural, zhao2023learning}
have received increasing attention thanks to the efficiency and scalability of deep networks. As garment deformation consists of both high-frequency wrinkles and low-frequency posed garment shape, it is challenging for neural networks to directly regress a single deformation field due to its spectral bias \citep{rahaman2019spectral, zhang2023parameter}. To this end, previous works mostly adopt a simplified two-stage decomposition by firstly regressing wrinkles relative to the un-posed garment template, then using
linear blend skinning (LBS) to obtain the low-frequency posed garment shape. However, due to the lack of explicit supervision on garment skinning, they either assume tight garments and directly skin from the closest body vertex \citep{santesteban2019learning, patel2020tailornet, santesteban2021self, santesteban2022snug}, or skin loose garments with virtual joints \citep{pan2022predicting, zhao2023learning}. Such unsupervised skinning can produce misaligned garment positions and undesired artifacts, which corrupts high-frequency signals. Consequently, these skinning-based methods can fail to recover high-fidelity wrinkles details, as illustrated in Figure \ref{fig:open}.

To tackle this issue, we present a novel \emph{skinning-free} method that decomposes high-low frequency modalities instead with two geometric attributes.
Specifically, we propose to directly estimate posed garment \textbf{vertex positions} instead of relying on garment skinning. As networks tend to prioritize learning low-frequency signals \citep{rahaman2019spectral}, we observe the model learns well with the overall posed shape but can generate over-smoothed garment geometry. To recover fine-grained wrinkles, we further estimate \textbf{vertex normals} that better capture local surface bending arising in wrinkles. In contrast to skinning-based methods, our method effectively decouples frequency components and enables \emph{explicit supervision} for both modalities, which avoids noisy skinning and produces high-quality wrinkles.

Motivated by the capability of visual perception in large vision models \citep{dosovitskiy2020image, caron2021emerging, oquab2023dinov2}, we propose to project 3D garments onto 2D image space aiming at improving perceptual quality of deformation results. 
Specifically, unlike previous works that mostly rely on manual UV partition for semantically-meaningful image representation \cite{lahner2018deepwrinkles, zhang2021deep}, we instead render both vertex attributes as multi-view texture images encoded in the garment \emph{canonical shape}, which efficiently scales to large collection of garments with diverse topologies and minimizes garment-body occlusion during deformation. In this way, we effectively model 3D garment deformation as a 2D image transfer task, as illustrated in Figure \ref{fig:open}. Finally, we fuse priors from transferred images of both frequency modalities to recover deformed 3D garments, which robustly generates plausible deformation for invisible areas introduced by rendering projection.

Our contributions can be summarized as follows. (\emph{i}) We propose a novel skinning-free pipeline for garment deformation with effective high-low frequency modalities decomposition, which avoids noisy garment skinning and facilitates fine-grained wrinkle regression. (\emph{ii}) We model 3D garment deformation via 2D image transfer, leveraging powerful vision models and scalable image representation to recover high-fidelity visual details for wrinkles. Extensive experiments show that our method noticeably improves deformation and perceptual quality over state-of-the-art learning-based methods.

\section{Related Works}

\textbf{Physics-based Methods. }To generate physically plausible garment animation, physics-based methods either rely on time-consuming simulators \citep{provot1995deformation, bouaziz2023projective, li2022diffcloth, yu2023diffclothai}, or optimize through physics-inspired losses \citep{bertiche2020pbns, santesteban2022snug, grigorev2023hood}. To ensure realism and accuracy, simulator parameters need to be fine-tuned for each garment instance, which can be laborious. Several works propose to estimate these parameters through differentiable simulation \citep{larionov2022estimating, li2023diffavatar} or neural networks \citep{yang2017learning, clyde2017modeling}, however, the estimation needs to be performed in a controlled setting with known external factors, which limits their applications. The challenge in data preprocessing thus restricts such method from scaling to diverse garment types. 

\noindent
\textbf{Learning-based Methods. }In contrast, learning-based methods  \citep{patel2020tailornet, santesteban2019learning, pan2022predicting, zhao2023learning, zhang2022motion, ma2022neural, chen2024gaps, wang2025fresa} have been developed to achieve superior efficiency and scalability. Pioneered by \citep{10.1145/344779.344862}, most works follow to estimate pose space deformation (PSD), namely they adopt LBS to obtain low-frequency posed garment shape, while predicting high-frequency wrinkles in the canonical garment space. Specifically, \citep{santesteban2019learning} directly regresses local vertex displacements using recurrent neural networks. \citep{patel2020tailornet} proposes to first use mixture models to construct bases of high-frequency deformations, then combine them with narrowed bandwidth kernels. \citep{zhang2022motion} leverages generative models to encode the feasible high-frequency latent space. Similar to our approach, \citep{lahner2018deepwrinkles, zhang2021deep} uses normal maps to model fine wrinkles. However, they require manually built UV maps and rely on LBS to generate initial normals, which we show in the ablation study that are sub-optimal. While the above works tackle tight garments and directly access body skinning weights, \citep{pan2022predicting, zhao2023learning} further extend to loose garments by predicting virtual garment joints to which garments are skinned. However, the prediction of virtual joints can not be explicitly supervised, which can lead to incorrect joint transformations. In summary, existing learning-based methods mostly suffer from noisy skinning that can not be directly supervised. Consequently, the skinning artifacts need to be jointly refined during wrinkle regression, which prevents them from estimating accurate wrinkles. In contrast, we present a \emph{skinning-free} approach, which effectively avoids noisy skinning and facilitates to generate more accurate wrinkles.

\begin{figure*}
  \setlength{\abovecaptionskip}{0.0cm}
  \setlength{\belowcaptionskip}{-0.cm}
    \centering  
    \includegraphics[width=0.95\textwidth]{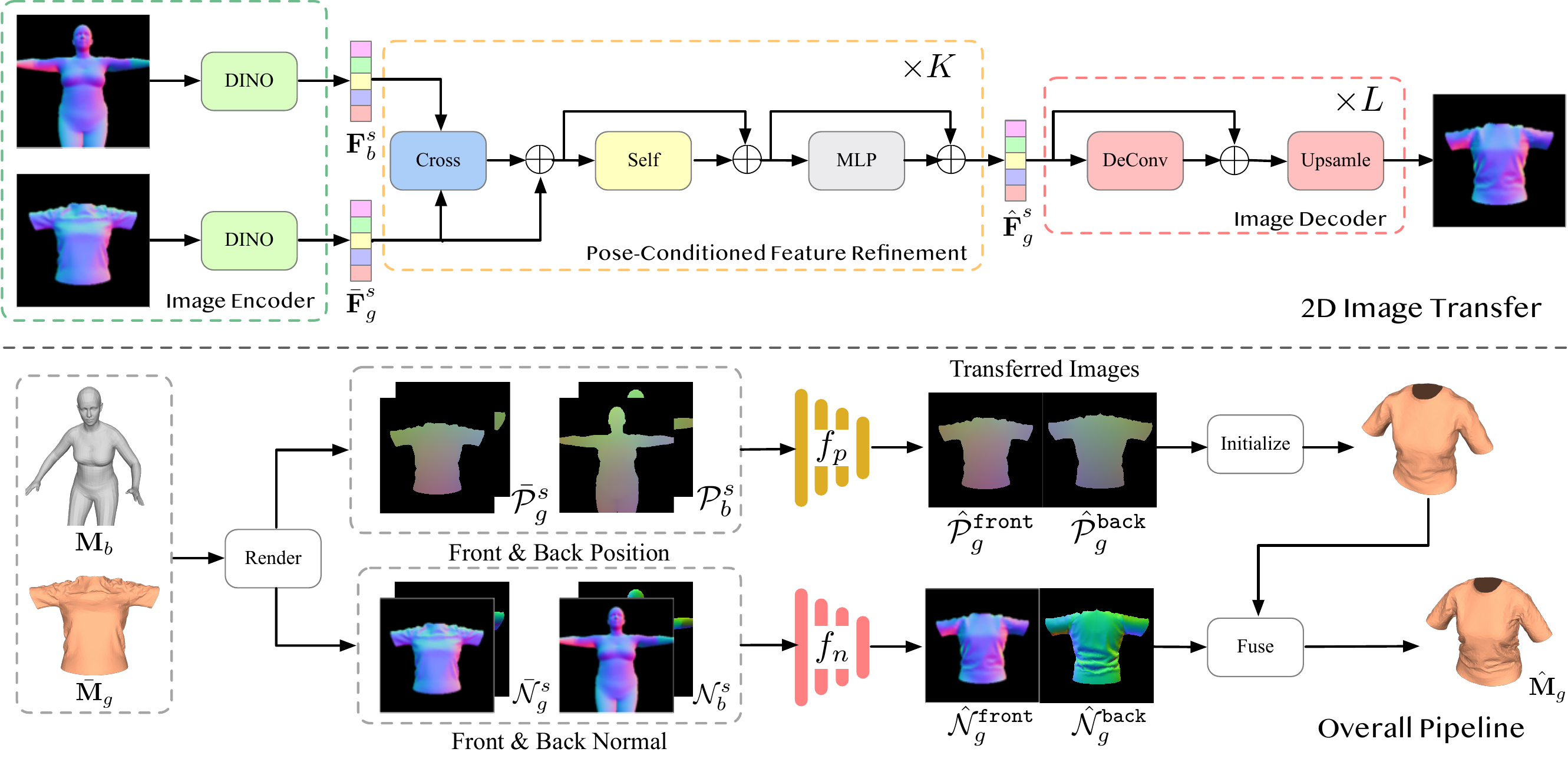}
    \caption{\textbf{Overview of our method.} Given the input garment template $\bar{\textbf{M}}_g$ and posed body mesh $\textbf{M}_b$, we first render position and normal images for the garment \{$\bar{\mathcal{P}}_g^s$, $\bar{\mathcal{N}}_g^s$\} and body \{${\mathcal{P}}_b^s$, ${\mathcal{N}}_b^s$\} from each view $s$, aiming to project the 3D garment onto the image space. Next, we transfer position images in $f_p(\cdot)$ and normal images in $f_n(\cdot)$, where the two networks have the same architecture as shown in the top row (taking front normal images as an example). Finally, we initialize the posed garment mesh from transferred position images $\hat{\mathcal{P}}_g^s$ and recover missing wrinkle details by fusing from normal images $\hat{\mathcal{N}}_g^s$ to obtain the deformed garment $\hat{\textbf{M}}_g$. "$\oplus$" denotes residual connection.}
    \label{fig:main}
    \vspace{-8pt}
\end{figure*}

\noindent
\textbf{Image-based 3D Representation. }In view of large-scale image datasets and effective pretrained image models, recent works propose to represent 3D geometry in the image space. Most existing works leverage UV mapping as the image representation, which have been widely applied in human pose estimation \citep{guler2018densepose}, avatar generation \citep{ma2021pixel, li2023animatable}, and scan registration \citep{guo2023diffusion}. However, they mostly rely on manual UV unwrapping to produce semantically meaningful islands, which requires expert knowledge and is laborious,  thus does not scale to large-scale collections. Alternatively,  \citep{lin2022collaborative, li2023animatable, xiu2022icon, xiu2023econ} propose to render multi-view images to automatically establish vertex-to-pixel correspondence. Specifically, \citep{xiu2022icon, xiu2023econ} learn to generate 3D clothed humans by integrating from estimated normal images. \citep{lin2022collaborative} encodes 3D character animation via ultra dense pose images. \citep{li2023animatable} designs Gaussian maps rendered from template human meshes to encode parameters for Gaussian splatting. Unlike all above works that consider only a single image source, we observe that accurate garment deformation requires effectively fusing \emph{multiple} image domains with \emph{mixed} frequency modalities, which is achieved via a novel pipeline as will be introduced below.

\section{Method}
To avoid skinning artifacts and produce more accurate wrinkles, we present a novel skinning-free pipeline, as shown in Figure \ref{fig:main}. Given a garment template mesh $\bar{\textbf{M}}_g=\{\bar{\mathbf{V}}_g, F_g\}$ where $\bar{\mathbf{V}}_g \in \mathbb{R}^{N \times 3}$ denotes vertex positions and $F_g \in \mathbb{Z}_{+}^{F \times 3}$ denotes triangle faces, we aim to estimate its deformed mesh $\hat{\textbf{M}}_g=\{\hat{\mathbf{V}}_g, F_g\}$ conditioned on the posed body mesh ${\textbf{M}}_b=\{{\mathbf{V}}_b, F_b\}$. Instead of directly regressing 3D vertex displacements, we propose to model garment deformation in 2D image space. Specifically, we first render RGB images $\bar{\mathcal{P}}_{g}^{s}$ and $\bar{\mathcal{N}}_{g}^s \in \mathbb{R}^{H \times W \times 3}$ that respectively encode vertex positions and normals of the garment template from each view $s \in \{\text{front},  \text{back}\}$. Similarly for the posed body mesh we render corresponding images $\{{\mathcal{P}}_{b}^{s}, {\mathcal{N}}_{b}^s\}$. We then develop an image transfer network to generate transferred images $\hat{\mathcal{P}}_{g}^{s}$ and $\hat{\mathcal{N}}_{g}^s$, which describe posed garment shape and wrinkles respectively (Section \ref{section:32}). Finally, we propose a multimodal fusion process to leverage priors of both modalities and optimize 3D deformed garment mesh from transferred images (Section \ref{section:33}).

\subsection{Image Rendering}
\label{section:31}
We represent 3D garment and body meshes as 2D images rendered from \emph{multiple views}, where the intensities of pixels encode the garment geometry, \emph{e.g.} positions or normals of vertices. Specifically, to generate such an image for a deformed mesh, we use its vertex positions and normals to color the corresponding vertices on the \emph{template mesh}, then render the results from both front and back views. In this way, for different deformations of the same mesh, the rendered pixel values will be different while the image silhouette \emph{remains the same}, as we always project the template mesh onto the image space. Taking the garment as an example, given vertex positions $\textbf{V}_g$ for a deformed mesh, we compute the corresponding vertex normals $\textbf{N}_g \in \mathbb{R}^{N \times 3}$ and linearly rescale their values to fit RGB colors, \emph{i.e.} within the range $[0,1]$. We then render the images from each view with a perspective camera of known transformation matrix as:
\begin{align}
    \mathcal{P}_g^s &= f_r^{s}(\text{RGB}(\textbf{V}_g); \bar{\mathbf{M}}_g) \\
    \mathcal{N}_g^s &= f_r^{s}(\text{RGB}(\textbf{N}_g); \bar{\mathbf{M}}_g) \;,
    \label{eq1}
\end{align}
where $\text{RGB}(\cdot)$ represents the linear rescaling function that maps positions or normals to RGB values, and $f_r^{s}(\cdot; \bar{\mathbf{M}}_g)$ represents the renderer function from view $s$ with the constant template $\bar{\mathbf{M}}_g$. Similarly, we can obtain images for the posed body as $\mathcal{P}_b^s$ and $\mathcal{N}_b^s$, where we use the same cameras as for the garment to capture only relevant body areas.

\noindent
\textbf{Discussion.} Our rendering configuration has two advantages compared to alternatives: 1) We can automatically establish vertex-to-pixel correspondence through perspective projection, thus do not require manually built UV parameterization \citep{lahner2018deepwrinkles} or sew patterns \citep{pietroni2022computational, li2023diffavatar}. The rendered images also retain the canonical garment shape, which facilitates the image feature extraction; 2) Instead of directly projecting the deformed mesh, we project the template mesh throughout rendering, \emph{which will not be influenced by self-occlusions during garment deformation.} Moreover, we use front and back views to efficiently capture most visible garment vertices, which also provide sufficient constraints to infer non-visible vertices at side views. After rendering all images, we can estimate 3D garment deformation via 2D image transfer as illustrated next section.

\subsection{2D Image Transfer}
\label{section:32}
As the garment geometry can be fully represented by the position and normal images, we formulate 3D garment deformation as an image transfer task, \emph{i.e.} we aim to transfer from the initial images $\{\bar{\mathcal{P}}_g^s, \bar{\mathcal{N}}_g^s\}$ representing the garment template to the target images $\{\mathcal{P}_g^s, \mathcal{N}_g^s\}$ representing the deformed garment, conditioned on the posed body images $\{{\mathcal{P}}_b^s, {\mathcal{N}}_b^s\}$ as:
\begin{equation}
    \hat{\mathcal{P}}_g^s = f_p(\bar{\mathcal{P}}_g^s, \mathcal{P}_b^s), \quad \quad \hat{\mathcal{N}}_g^s = f_n(\bar{\mathcal{N}}_g^s, \mathcal{N}_b^s) \;,
\end{equation}
where $\hat{\mathcal{P}}_g^s$ and $\hat{\mathcal{N}}_g^s$ are the estimated posed position and normal images, respectively. The position transfer network $f_p(\cdot)$ and normal transfer network $f_n(\cdot)$ have the same architecture (as in Figure \ref{fig:main}). Each network contains three consecutive modules: (\emph{i}) an image feature encoder that extracts visual features of garment and body geometry, (\emph{ii}) a pose-conditioned feature refinement module that injects pose condition and models fine-grained body-garment interaction, and (\emph{iii}) an image decoder that decodes the transferred images. We will introduce each module in below.

\noindent
\textbf{Image Feature Encoder.} We forward each image input to a pre-trained vision transformer DINO \citep{caron2021emerging} to encode patch-wise tokens of image features $\bar{\textbf{F}}_g^s, {\textbf{F}}_b^s \in \mathbb{R}^{M \times D}$ for garment and body respectively, where $M$ represents the number of tokens and $D$ represents the feature dimension. Compared with other encoders like ImageNet-pretrained ResNet \citep{he2016deep}, DINO can effectively encode detailed structural and visual information through attention on salient image contents, which is beneficial for generating fine wrinkles. We further show its efficacy in Section \ref{sec:ab}.

\noindent
\textbf{Pose-Conditioned Feature Refinement.} We refine image features to introduce pose priors in $K$ transformer blocks. In each block, motivated by \citep{wang2023interacting}, we model body-garment interaction by first computing the multi-head cross-attention \citep{vaswani2017attention} between the garment feature $\bar{\textbf{F}}_g^s$ and the body feature ${\textbf{F}}_b^s$ to generate the pose-conditioned garment feature. In contrast to the traditional skinning process that computes the skinning weights with respect to sparse joints, we learn to model the \emph{dense} correlation between image patches, which can capture fine-grained body-garment interaction. Furthermore, we follow the vanilla transformer structure \citep{vaswani2017attention} and continue to forward the feature into a self-attention layer followed by a multi-layer perceptron (MLP) to generate the refined garment feature $\hat{\textbf{F}}_g^s$.

\noindent
\textbf{Image Decoder.} Finally, $\hat{\textbf{F}}_g^s$ is rearranged spatially to form a 3D tensor corresponding to the 2D image feature map and forwarded to an image decoder to generate transferred images. The image decoder consists of residual blocks of 2D convolution layers, followed by transposed convolution layers to upsample the spatial resolution.

\noindent
\textbf{Training Objectives. }We train each network using the masked L1 loss as:
\begin{align}
    \mathcal{L}_p &= {\textstyle\sum}_s|| \bar{\mathcal{S}}_g^s \odot \hat{\mathcal{P}}_g^s - \bar{\mathcal{S}}_g^s \odot {\mathcal{P}}_g^s ||_1 \\
    \mathcal{L}_n &= {\textstyle\sum}_s|| \bar{\mathcal{S}}_g^s \odot \hat{\mathcal{N}}_g^s - \bar{\mathcal{S}}_g^s \odot {\mathcal{N}}_g^s ||_1 \;,
\end{align}
where ${\mathcal{P}}_g^s$ and ${\mathcal{N}}_g^s$ represent ground truth position and normal images, $\bar{\mathcal{S}}_g^s$ represents the silhouette of the garment template to mask for valid pixels, and $\odot$ represents pixel-wise multiplication. Note that we independently model each modality regardless of their consistency constraints, as we observe that mixing frequency components during image transfer leads to inferior accuracy (as compared in Section \ref{sec:ab}). Alternatively, we opt to fuse position-normal correlation via explicitly optimization.

\subsection{3D Multimodal Fusion}
\label{section:33}
While we can obtain vertex positions of the deformed garment solely from the position images $\hat{\mathcal{P}}_g^s$ of both views, we observe two major issues of such an approach: (\emph{i}) since the high frequency wrinkle details are often reflected by relatively small position changes, it is hard for the position transfer network $f_p(\cdot)$ to capture such subtleties, thus leading to an over-smoothed mesh, (\emph{ii}) although images from front and back views cover most of the garment, the positions of non-visible vertices at side views can not be directly obtained from these images. In this section, we propose a multimodal fusion process to address both issues. The key idea is to incorporate high-frequency wrinkle details recorded by the normal images to refine the over-smoothed mesh initialized from the position images, while using the edge and surface priors to constrain the non-visible vertices. Specifically, we aim to optimize the deformed vertex positions $\textbf{V}_g^{\star}$ that aligns with both image observations as:
\begin{equation}
    \textbf{V}_g^{\star} \!=\! \arg \min_{\hat{\textbf{V}}_g} {\textstyle\sum}_s(||f_r^s(\hat{\textbf{V}}_g) - \hat{\mathcal{P}}_g^s|| + ||f_r^s(\hat{\textbf{N}}_g) - \hat{\mathcal{N}}_g^s||) \;,
    \label{eq4}
\end{equation}
where $f_r^s(\cdot)$ is the renderer function in Eq.(\ref{eq1}) that omits constants and $\hat{\textbf{N}}_g$ are vertex normals computed from vertex positions $\hat{\textbf{V}}_g$. The optimization includes two stages where garment mesh is firstly initialized from position images and then refined with normal images to recover fine wrinkles.

\noindent
\textbf{Vertex Position Initialization. }We initialize vertex positions from position images based on their visibility under the perspective projection. For visible vertices, we simply interpolate their corresponding pixel values in the transferred position images to initalize their positions. For non-visible vertices especially at side views, we initialize them by linearly interpolating from the closest front and back visible vertex pairs. To correct the linear interpolation and ensure a smooth boundary between two types of vertices, we smooth the results by minimizing the edge length loss $\mathcal{L}_e$ as:
\begin{equation}
    \mathcal{L}_{e} \!=\!\frac{1}{|\mathcal{E}|} {\textstyle\sum}_{\{i, j\} \in \mathcal{E}}(\|\hat{\textbf{V}}_g[i] \!-\! \hat{\textbf{V}}_g[j]\| \!-\! \|\bar{\textbf{V}}_g[i] \!-\! \bar{\textbf{V}}_g[j]\| )^2 \;,
    \label{eq5}
\end{equation} 
where $\mathcal{E}$ represents the index set of all edges defined by the garment faces $F_g$, $\hat{\textbf{V}}_g$ and $\bar{\textbf{V}}_g$ represent estimated and template mesh vertices, respectively. Moreover, we impose a regularization loss $\mathcal{L}_{rv}$ to penalize the $L_2$ distance on displacements of visible vertices, in order to align with the position images. The overall loss for this stage can be summarized as $\mathcal{L}_e + \lambda_{rv} \mathcal{L}_{rv}$, with loss weight $\lambda_{rv}$.

\noindent
\textbf{Vertex Normal Fusion. }To amend high-frequency wrinkle details on the position-initialized vertices, we fuse normal predictions onto them by minimizing the normal rendering loss $\mathcal{L}_r$ defined as the second term in Eq.(\ref{eq4}), and then smooth the results using a normal consistency loss $\mathcal{L}_{rn}$ as:
\begin{equation}
  \mathcal{L}_{rn} = \frac{1}{|\mathcal{E}|} {\textstyle\sum}_{\{i, j\} \in \mathcal{E}}(1 - \hat{\mathbf{N}}_g[i]^T\hat{\mathbf{N}}_g[j]) \;.
\end{equation}
Similar to the initialization stage, we impose the edge length loss $\mathcal{L}_e$ to penalize irregular rim contours and include the vertex displacement regularization $\mathcal{L}_{rv}$ on all vertices. Finally, to penalize garment-body collision, we impose a collision loss by penalize the penetration distance as:
\begin{equation}
    \mathcal{L}_c = \frac{1}{N}{\textstyle\sum}_i\max(0, -\text{SDF}(\hat{\textbf{V}}_g[i],\textbf{M}_b))\;,
\end{equation}
where $\text{SDF}(\cdot)$ represents the vertex-to-mesh signed distance. The overall optimization objectives for normal fusion can be summarized as:
\begin{equation}
    \mathcal{L} = \mathcal{L}_r + \lambda_{rn}\mathcal{L}_{rn} + \lambda_{e}\mathcal{L}_e + \lambda_{rv} \mathcal{L}_{rv} + \lambda_{c}\mathcal{L}_c \;,
\end{equation}
where $\lambda_{rn}, \lambda_{e}, \lambda_{rv}, \lambda_{c}$ balance the weights of losses.

\section{Experiments}

\subsection{Experiment Setup}
\textbf{Datasets:} We evaluate our method and baselines on three benchmarks: 1) \textbf{VTO} dataset \citep{santesteban2019learning}containing two types of garments "t-shirt" and "dress" Each garment is draped onto an SMPL \citep{SMPL:2015} human body using ground truth deformations simulated in the ARCSim \citep{narain2014arcsim} simulator. We follow \citep{pan2022predicting} to use 4 clips (\texttt{01\_01}, \texttt{111\_02}, \texttt{55\_27} and \texttt{91\_36}) of medium body shape ($\bm{\beta}$ = 0) and unseen poses for testing and the remaining 49 clips for training. 2) \textbf{TailorNet} dataset \citep{patel2020tailornet}. Since VTO only contains upper garments, we further adopt TailorNet to test on lower garments "pants" and "skirt". We follow \citep{pan2022predicting} to use the medium body shape and garment style ($\bm{\beta}$ = 0, $\bm{\gamma}$ = 0) split, and adopt 2 clips (\texttt{005}, \texttt{010}) of unseen poses for testing and the remaining 16 clips for training. 3) \textbf{CLOTH3D} dataset \citep{bertiche2020cloth3d}. For the first two datasets, we test on a single body and garment shape \emph{only to ensure a fair comparison} with the commonly used evaluation standard in \cite{pan2022predicting, zhao2023learning}. However, our method is not limited to this set up. To verify this, we further demonstrates results on unseen dress garments from CLOTH3D for generalization evaluation (Section \ref{sec:cloth3d}).

\noindent
\textbf{Metrics.} Following \citep{pan2022predicting, zhao2023learning}, we evaluate all methods on three metrics: Root Mean Squared Error (RMSE), Hausdorff distance \citep{attouch1991topology}, and spatio-temporal edge difference (STED) \citep{vasa2010perception} between predicted and ground truth meshes. Specifically, RMSE and Hausdorff distance assess the prediction accuracy in, while STED evaluates the perceptual quality of deformation by measuring relative edge differences in each test clip. 

\noindent
\textbf{Implementation Details.} We render all images at $256 \times 256$ pixels using differentiable renderer from Nvdiffrast \citep{Laine2020diffrast} and normalize deformed garment and posed body vertices with the global rotation and translation from the human pose. For the image transfer network, we fine-tune the last two layers of the DINO encoder, along with $K = 4$ transformer blocks for the feature refinement module. We train the model using the Adam \citep{kingma2014adam} optimizer for 100K iterations, and set the learning rate to $1 \times 10^{-4}$. For multimodal fusion, we use the same optimizer with a learning rate of $1 \times 10^{-3}$ and optimize for 100 steps in both the initialization and normal fusion stages. For the loss weights, we set $\lambda_{rv} = 0.02$, $\lambda_e = \lambda_c = 100$, and $\lambda_{rn} = 0.001$ on t-shirt and $0.01$ on other garments based on their scales. We include the detailed model architecture and inference time comparison in supplement materials.

\subsection{Results on Garment Deformation}
Following \citep{pan2022predicting, zhao2023learning}, we report all metrics by training and testing on each garment instance to ensure a fair comparison. In Table \ref{table1}, we present the results on the VTO dataset, where metrics for baselines \citep{patel2020tailornet, santesteban2019learning, pan2022predicting, zhao2023learning} are borrowed from \citep{pan2022predicting, zhao2023learning} where we followed the same test configuration. For \citep{santesteban2021self}, we use its official weights and evaluate the results on our test split. We observe that our method achieves the best accuracy against all skinning-based methods \citep{patel2020tailornet, santesteban2019learning, pan2022predicting, zhao2023learning} thanks to the proposed skinning-free pipeline that avoids artifacts of noisy garment skinning. In particular, we outperform \citep{pan2022predicting, zhao2023learning} on loose garments, without the need to estimate additional virtual joints or anchors to facilitate garment skinning. Moreover, thanks to the capability of perceptual learning in image models \citep{amir2021deep}, our method achieves improved perceptual quality (lower STED values) on deformed garments.

\begin{table}[!tb]
\centering

\caption{\textbf{Quantitative comparison on the VTO Dataset.} Best results are highlighted in \textbf{bold}. Our method achieves superior deformation accuracy and perceptual quality compared to state-of-the-art skinning based methods \citep{patel2020tailornet, santesteban2019learning, pan2022predicting, zhao2023learning, santesteban2021self} on both tight and loose garments.}
\vspace{-0.2cm}
\label{tab:garment_deformation}
\resizebox{0.48\textwidth}{!}{
\small
\setlength{\tabcolsep}{2pt}
\begin{tabular}{@{}lcccccc@{}}
\toprule
\multirow{2}{*}{Methods} & \multicolumn{3}{c}{Dress} & \multicolumn{3}{c}{T-shirt} \\ \cmidrule(l){2-4} \cmidrule(l){5-7} 
                 & RMSE $\downarrow$ & Hausdorff $\downarrow$ & STED $\downarrow$ & RMSE $\downarrow$ & Hausdorff $\downarrow$ & STED $\downarrow$ \\ \midrule
TailorNet \citep{patel2020tailornet}   & 22.95 & 76.80 & 0.0757 & 9.90 & 27.02 & 0.0418  \\
Santesteban \citep{santesteban2019learning} &  20.96 & 87.01 & 0.0745 & 10.25 & 29.56 & 0.0449 \\
Santesteban \citep{santesteban2021self} & 21.07 & 87.98 & 0.0620 & 9.97 & 25.64 & 0.0335 \\
VirtualBones \citep{pan2022predicting} & 19.91 & 83.39 & 0.0722 & 10.52 & 31.51 & 0.0452 \\
AnchorDEF  \citep{zhao2023learning}      & 16.05 & 74.20 & 0.0493 & 6.25 & 26.31 & 0.0262 \\
Ours & \textbf{13.40} & \textbf{61.73} & \textbf{0.0407} & \textbf{4.66} & \textbf{20.89} & \textbf{0.0205} \\
\bottomrule
\end{tabular}
}
\label{table1}
\vspace{-5pt}
\end{table}

We further report quantitative comparison results in Table \ref{table2}. For pants, we borrow the metrics results reported by \citep{pan2022predicting} which has the same set-up with us. Since \citep{santesteban2019learning, pan2022predicting} do not release the training code or pretrained weights for the skirt split, we are unable to evaluate their results and thus only compare with \citep{patel2020tailornet} and retrain with their official code on the single garment set-up. We observe that our method consistently outperforms the skinning-based method \citep{patel2020tailornet} when applied to lower garments.

\begin{table}[!tb]
\centering
\caption{\textbf{Quantitative comparison on the TailorNet Dataset.} Best results are highlighted in \textbf{bold}
and and inapplicable results are marked with "-". Our method consistently generates more accurate results than baseline
methods on lower garments.}
\vspace{-0.2cm}
\resizebox{0.48\textwidth}{!}{
\small
\setlength{\tabcolsep}{2pt}
\begin{tabular}{@{}lcccccc@{}}
\toprule
\multirow{2}{*}{Methods} & \multicolumn{3}{c}{Pants} & \multicolumn{3}{c}{Skirt} \\ \cmidrule(l){2-4} \cmidrule(l){5-7} 
                 & RMSE $\downarrow$ & Hausdorff $\downarrow$ & STED $\downarrow$ & RMSE $\downarrow$ & Hausdorff $\downarrow$ & STED $\downarrow$ \\ \midrule
TailorNet \citep{patel2020tailornet} & 4.84 & 14.46 & 0.0127   
 & 7.76 & 16.28 & 0.0162 \\
Santesteban \citep{santesteban2019learning} & 4.91 & 14.87 & 0.0129 & - & - & - \\
VirtualBones \citep{pan2022predicting} & 4.76 & 18.75 & 0.0166 & - & - & -  \\
Ours & \textbf{4.03} & 
\textbf{13.55} & \textbf{0.0114} & \textbf{5.38} & \textbf{14.06} & \textbf{0.0150}\\
\bottomrule
\end{tabular}
}
\label{table2}
\vspace{-0.5cm}
\end{table}

\begin{figure*}[!tb]
    \centering
    \includegraphics[width=.9\textwidth]{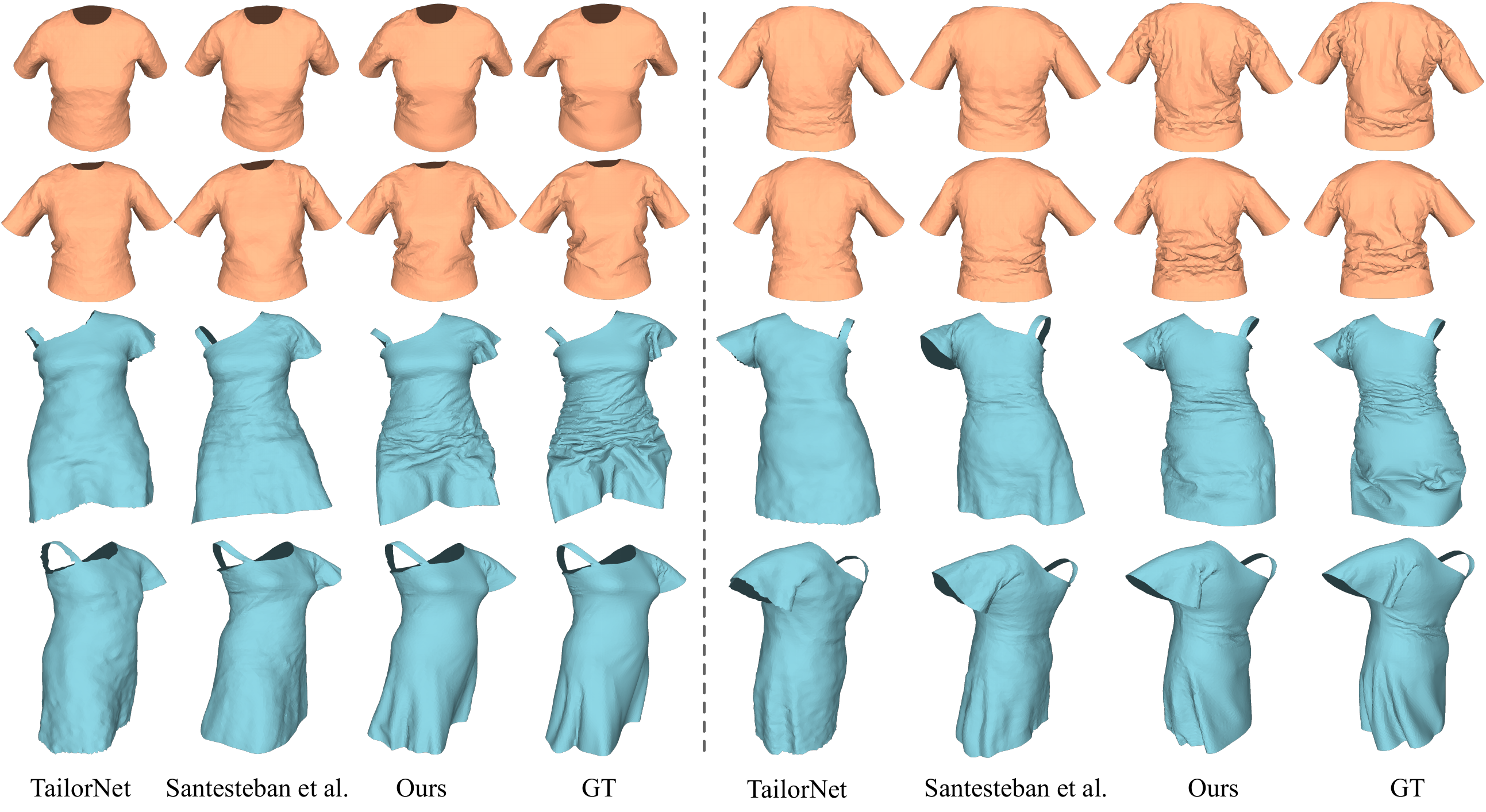}
    \vspace{-0.2cm}
    \caption{\textbf{Results on VTO dataset. }We produce more accurate wrinkles and folds than skinning-based methods \citep{patel2020tailornet, santesteban2021self}.}
    \label{fig:quali1}
    \vspace{-0.2cm}
\end{figure*}

\begin{figure*}[!tb]
    \centering
\includegraphics[width=.9\textwidth]{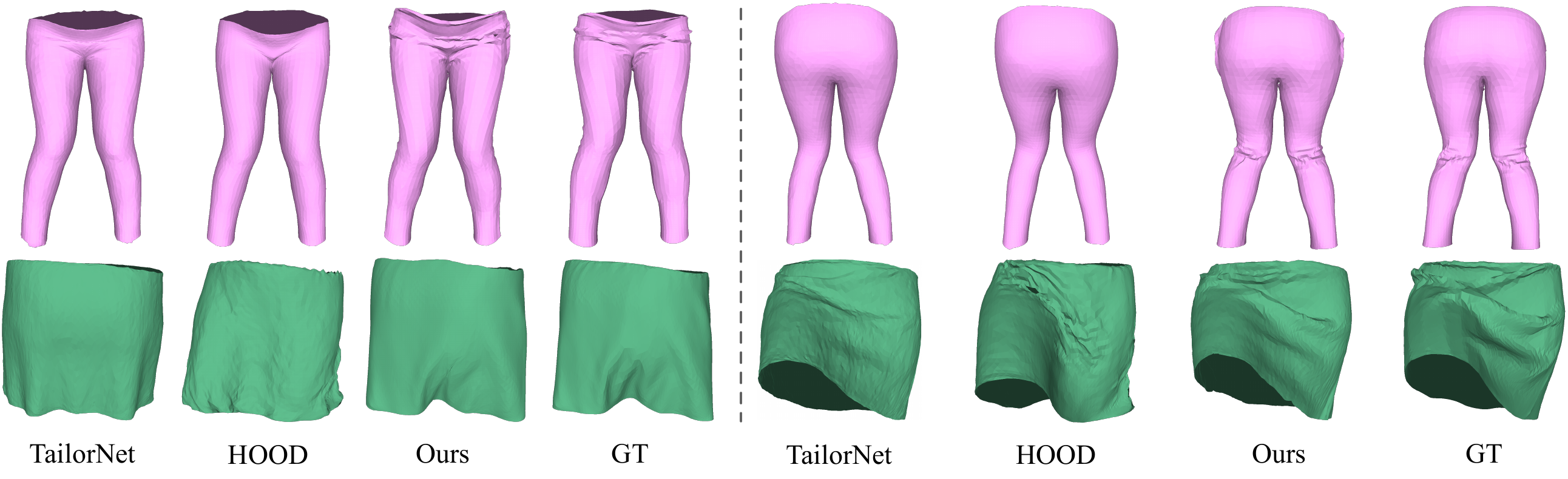}
    \vspace{-0.2cm}
    \caption{\textbf{Results on TailorNet dataset. }Our method consistently produces more accurate deformations on lower garments than \citep{patel2020tailornet, grigorev2023hood}.}
    \label{quali2}
    \vspace{-0.2cm}
\end{figure*}

\begin{figure*}[htp!]
    \centering
\includegraphics[width=0.9\textwidth]{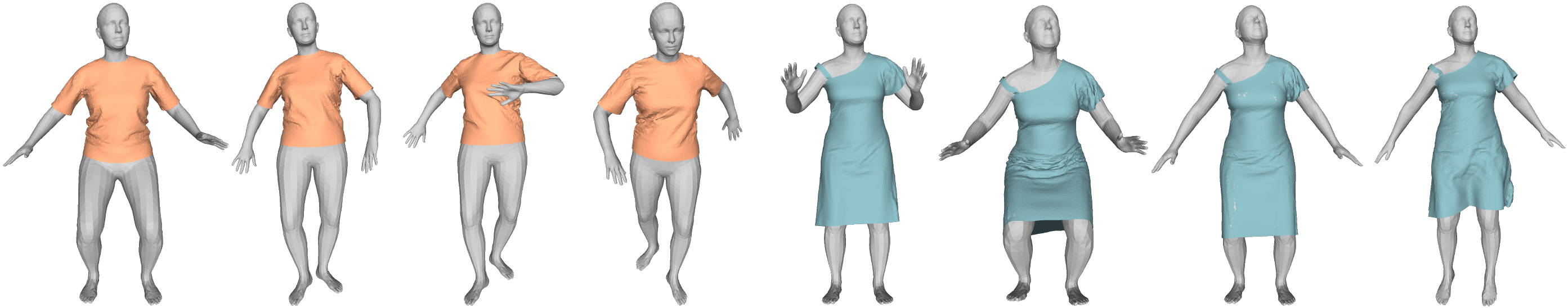}
    \vspace{-0.2cm}
    \caption{\textbf{Qualitative results with human motions. }Our method generates accurate and plausible garment deformations for a sequence of unseen human poses. Moreover, the deformed garments are temporally consistent. We show more results in the supplementary video.}
    \vspace{-0.5cm}
    \label{quali:temp}
\end{figure*}

\noindent
\textbf{Qualitative results.} As shown in in Figures \ref{fig:quali1} and \ref{quali2}, our method can generate 3D deformed garments with finer wrinkle details and more accurate fold patterns. In comparison, the skinning-based baselines \citep{patel2020tailornet, santesteban2021self} can not recover accurate wrinkles in challenging cases due to the corrupted high-frequency signals, thus generating over-smoothed geometries. For completeness, we further include a comparison with physics-based methods
\citep{grigorev2023hood}, which requires fine-tuning physical parameters for each garment to ensure accurate deformation. However, the fine-tuning process is non-trivial with only inputs of garment geometry, thus are not directly comparable to our method. We show that directly using their default physical parameters can lead to unrealistic behaviors, \emph{e.g.} over-stretched skirts, as these physical parameters often have a global effect on the overall deformation. In contrast, learning-based methods like us ensures a more robust results by only estimating local vertex displacements. Moreover, we show in Figure \ref{quali:temp} that accurate estimation produces results with reduced penetration, which align well with the underlying body motions. Finally, we show in the supplementary video that accurate results also lead to temporal consistency, although we use only a single pose as condition. We attribute this to the continuous nature of the neural networks on continuous inputs, which is similarly observed in previous works \cite{patel2020tailornet} with the same setup.

\subsection{Ablation Study}
\label{sec:ab}
\textbf{Skinning-free Approach.} To show the efficacy of the skinning-free approach, we compare in Table \ref{tbc} with two LBS-based variants (both using skinning weights from nearest body vertices) equipped with our image transfer modules: (\emph{i}) we supervise with GT transferred images in the canonical space, which are generated via inverse LBS (IT + LBS), and (\emph{ii}) we refine from input images of LBS re-posed garments (LBS + IT), analog to \citep{lahner2018deepwrinkles, zhang2021deep}. We observe that both variants produce inferior results, as inaccurate skinning can produce noisy artifacts, which corrupt high-frequency signals in either GT or input images and thus hindering learning correct wrinkle patterns.

\begin{table}[!tb]
\centering
\caption{\textbf{Effects of the skin-free approach.} We show that mixing linear blend skinning with image transfer (IT) leads to inferior performance than our full skinning-free approach.}
\vspace{-0.2cm}
\resizebox{0.48\textwidth}{!}{
\small
\setlength{\tabcolsep}{2pt}
\begin{tabular}{@{}lcccccc@{}}
\toprule
\multirow{2}{*}{Methods} & \multicolumn{3}{c}{Dress} & \multicolumn{3}{c}{T-shirt} \\ \cmidrule(l){2-4} \cmidrule(l){5-7} 
                 & RMSE $\downarrow$ & Hausdorff $\downarrow$ & STED $\downarrow$ & RMSE $\downarrow$ & Hausdorff $\downarrow$ & STED $\downarrow$ \\ \midrule
IT + LBS & 18.50 & 78.25 & 0.0625 & 6.85 & 25.30 & 0.0295 \\
LBS + IT & 15.75 & 68.40 & 0.0493 & 6.01 & 24.15 & 0.0242 \\
Ours (Skin-free) & \textbf{13.40} & \textbf{61.73} & \textbf{0.0407} & \textbf{4.66} & \textbf{20.89} & \textbf{0.0205} \\
\bottomrule
\end{tabular}
}
\label{tbc}
\vspace{-0.5cm}
\end{table}

\begin{figure}[htp!]
    \centering
\includegraphics[width=0.37\textwidth]{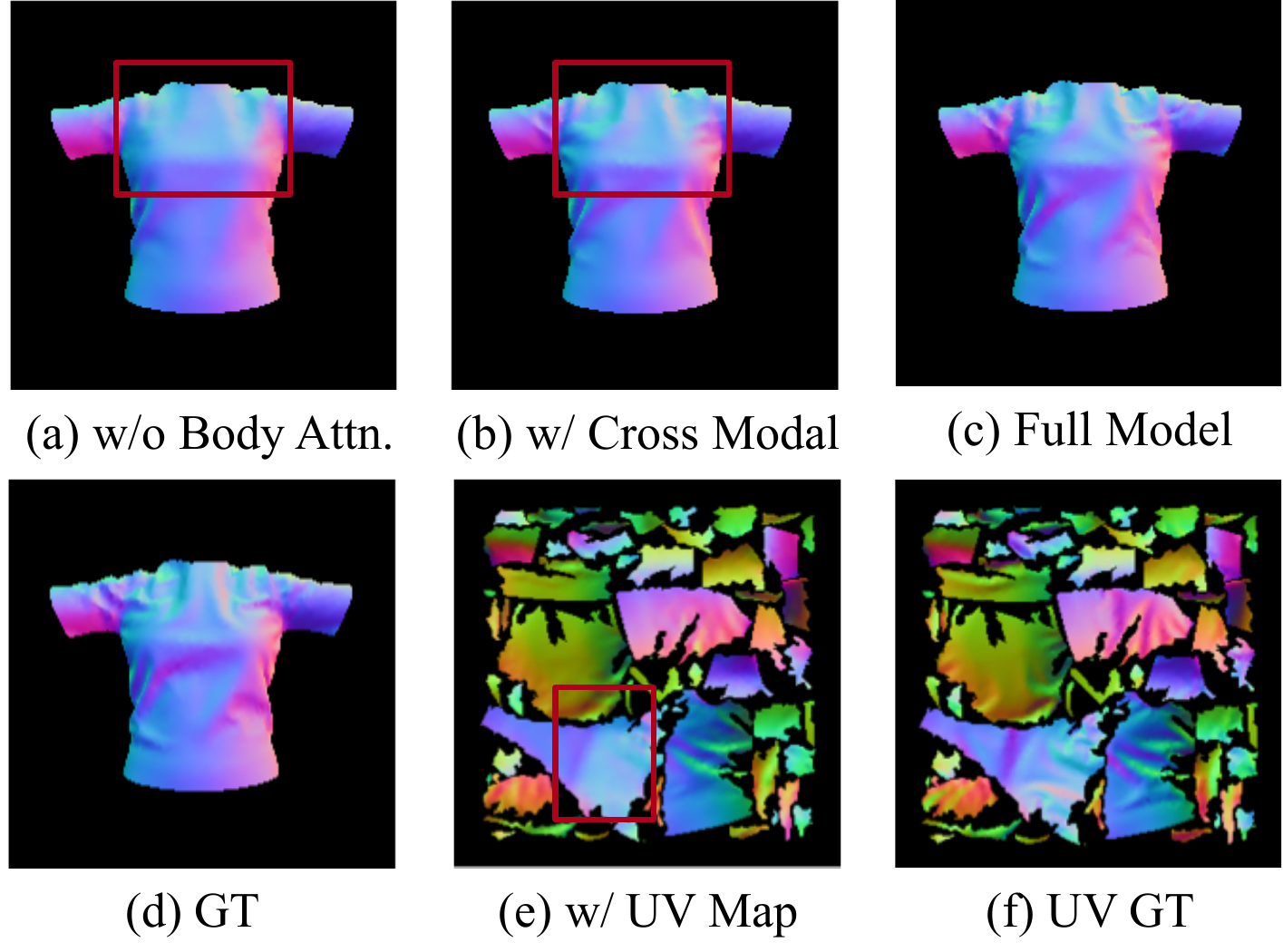}
    \vspace{-0.2cm}
    \caption{\textbf{Comparison of design variants for image transfer modules.} We show that design variants in the network architecture (a) and (b) produce smoother results, while the full model (c) can generate fine wrinkles that are closer to the GT (d). In addition, we show that using automatically generated UV maps (e) results in complex islands that are not beneficial for wrinkle estimation.}
    \label{ab1}
    \vspace{-0.4cm}
\end{figure}

\noindent
\textbf{Study on Image Transfer (IT).} To verify the effects of key modules in the image transfer network, we compare several alternatives to the current designs: (\emph{i}) replacing body-garment cross-attention with simply adding the body and garment features (w/o Body Attn.) (\emph{ii}) replacing the DINO encoder with ResNet-50 (ResNet Encoder), and (\emph{iii}) adding cross-attention between two modalities (w/ Corss Modal), as illustrated in Figure \ref{ab1}. We find both (\emph{i}) and (\emph{ii}) lead to smoothed garment geometry, showing the efficacy of the body-garment cross-attention for modeling fine-grained body-garment interaction, as well as the DINO encoder for extracting detailed garment structural information. Moreover, we empirically observe that mixing frequency signals like (\emph{iii}) results in inferior accuracy, thus we choose to separately tackle each modality.

In addition, we compare with the image representation of automatically generated UV maps via xatlas in Figure \ref{ab1} and Table \ref{tab:table1}. Since such images contain a large number of islands and therefore destroys garment shape priors, which is not conducive to the pretrained encoders and thus does not benefit wrinkle estimation. While manual UV can potentially solve this issue, it does not scales to a large collection of garments. Finally, we quantitatively compare in Table \ref{tab:table1} with two more variants, which uses 3D MLP and graph convolution network (GCN) to directly regress 3D vertex positions and normals.
Due to limited capacity of simple 3D networks, it produces inferior results than the proposed image transfer module.

\begin{figure}[!tb]
    \centering
\includegraphics[width=0.48\textwidth]{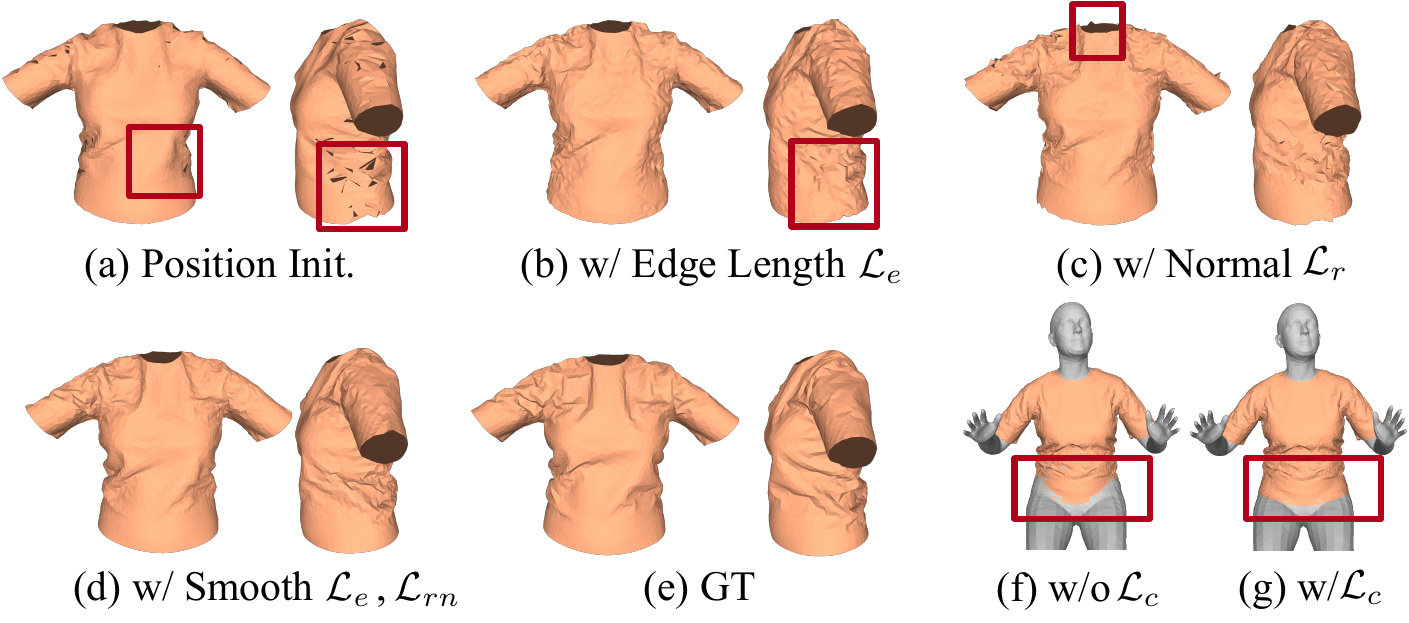}
    \vspace{-0.7cm}
    \caption{\textbf{Illustration of intermediate fusion results.} We show optimization results after adding corresponding losses. By fusing both modalities, we produce more accurate deformation with higher perceptual quality. Moreover, the edge length and normal consistency help to constrain non-visible vertices and resolve artifacts at mesh rims.}
    \label{ab2}
    \vspace{-0.5cm}
\end{figure}

\noindent
\textbf{Fusion Losses.} In Table \ref{tab:table2} and Figure \ref{ab2}, we show the effect of each loss during fusion optimization. The initial mesh (a) interpolated from the position images does not contain enough high-frequency wrinkle details, and non-visible vertices at side views can not be constrained, which leads to large RMSE error. By enforcing edge length consistency, we repair non-visible vertices as in (b). Jointly with (a) and (b), we obtain a reasonably good initialization that allows normal fusion to be feasibly achieved with few optimization steps. Moreover, we recover more accurate wrinkles after fusing from normal images (c). However, with only the normal loss, we observe artifacts at mesh rims due to under-constrained objectives. We thus further refine the results by penalizing irregular boundary edges and abrupt normal changes, which generate smoothed results as in (d). Finally, we show that the collision loss $\mathcal{L}_c$ effectively reduces body-garment collision as shown in (g).

\begin{table}[htp!]
    \centering
    \vspace{-0.2cm}
    \begin{minipage}{0.49\linewidth}
        \centering
        \caption{\textbf{Variants with IT.}}
        \vspace{-0.2cm}
        \resizebox{\linewidth}{!}{
        \small
        \setlength{\tabcolsep}{2pt}
        \begin{tabular}{@{}lccc@{}}
            \toprule
             Methods & RMSE$\downarrow$ & Hausdorff$\downarrow$ & STED $\downarrow$ \\ \midrule
            3D MLP & 8.39 & 24.77 & 0.0310 \\
            3D GCN & 7.95 & 24.25 & 0.0337  \\
            Auto-UV & 6.43 & 23.38 & 0.0277 \\ \midrule
            \textbf{Full Model} & \textbf{4.66} & \textbf{20.89} & \textbf{0.0205}  \\ \bottomrule
        \end{tabular}
        }
        \label{tab:table1}
    \end{minipage}
    \hfill
    \begin{minipage}{0.49\linewidth}
        \centering
        \caption{\textbf{Study on fusion losses.}}
        \vspace{-0.2cm}
        \resizebox{\linewidth}{!}{
        \small
        \setlength{\tabcolsep}{2pt}
        \begin{tabular}{@{}lccc@{}}
            \toprule
             Methods & RMSE$\downarrow$ & Hausdorff$\downarrow$ & STED $\downarrow$ \\ \midrule
            Init. (no normal) & 9.98 & 30.04 & 0.0771  \\
            Init. + $\mathcal{L}_e$ & 5.52 & 23.45 &  0.0525 \\
            Init. + $\mathcal{L}_e$ +  $\mathcal{L}_{r}$ & 4.96 & 22.23 & 0.0238  \\ \midrule
            \textbf{Full Model} & \textbf{4.66} & \textbf{20.89} & \textbf{0.0205}  \\ \bottomrule
            
        \end{tabular}
        }
        \label{tab:table2}
    \end{minipage}
    \vspace{-0.2cm}
\end{table}

\subsection{Generalization Analysis}
\label{sec:cloth3d}
\begin{figure}[!tb]
    \centering
\includegraphics[width=0.48\textwidth]{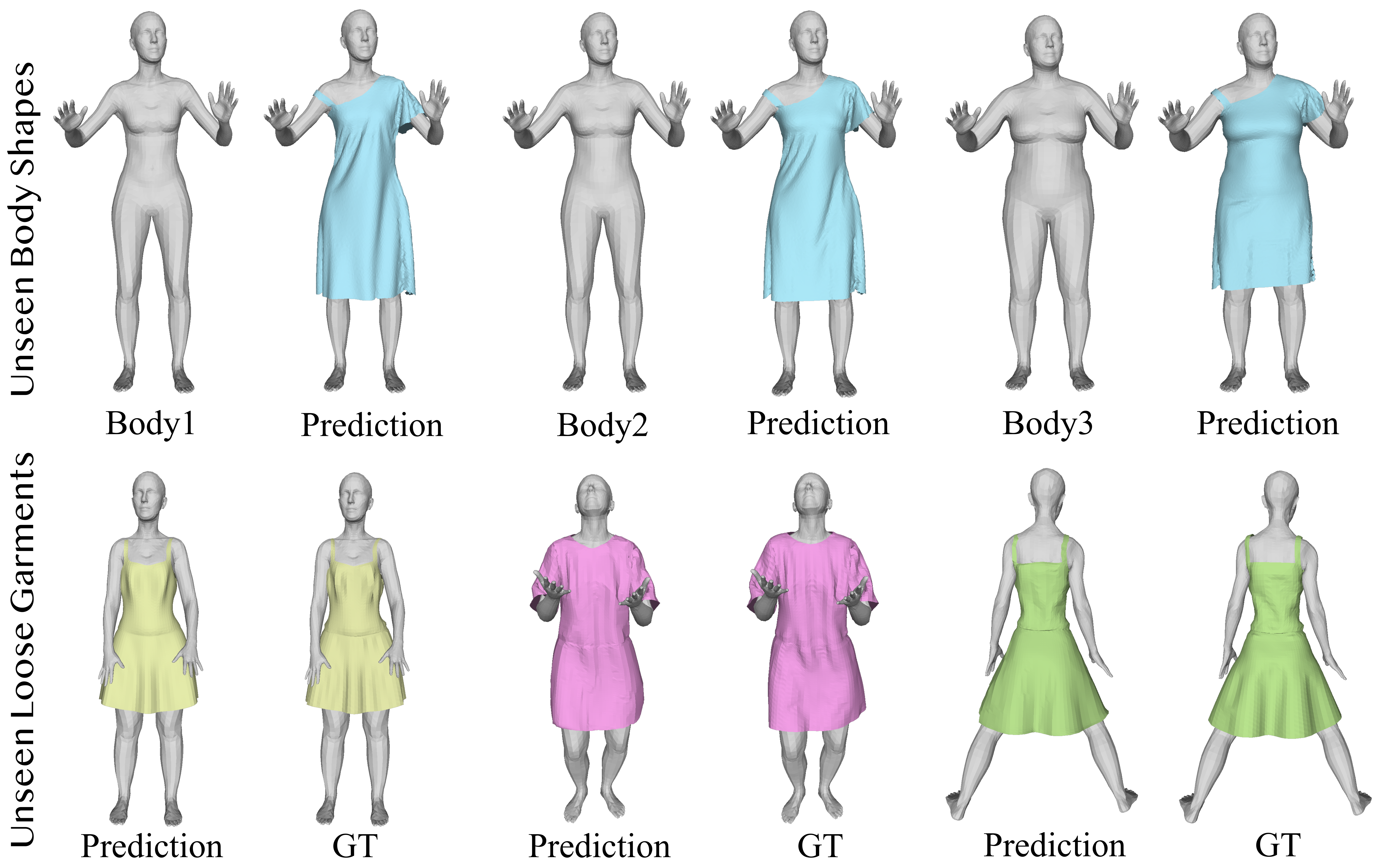}
\vspace{-0.7cm}    \caption{\textbf{Generalization results on unseen body shapes and garment types. }We show our method can be jointly trained on multiple body shapes and garment types to achieve effective generalization, demonstrating the scalability of the approach.}
    \label{fig:gene_cn}
    \vspace{-0.6cm}
\end{figure}

While baseline methods \citep{santesteban2019learning, santesteban2021self, pan2022predicting, zhao2023learning} are often evaluated for single garment set-up, our method is not limited to this configuration and in fact is well-suited for training across multiple garments thanks to several designs: (\emph{i}) the use of pretrained DINO encoder that is capable of extracting detailed semantic features for various garments, (\emph{ii}) the image transfer approach that is agnostic to garment topologies, and (iii) the use of front and back view projections to establish image representations for garments, which will not be scalable for manual UV parameterization on a large collection of garments. To verify the generalizability of our method, we further jointly train on 50 dress garments on the CLOTH3D \citep{bertiche2020cloth3d} dataset, and show the results in Figure \ref{fig:gene_cn}. By training across multiple garments, our method can effectively generalize to unseen garment shapes, with the skinning-free method particularly beneficial for tackling loose garments. Furthermore, we show that our method can be extended to jointly train on multiple body shapes, which enables generalization to unseen body shapes and producing different plausible deformations.

\section{Discussion}

\textbf{Limitation.} Although our method succeeds in generating fine-grained garment wrinkles, the geometric fidelity of the deformation are still
bounded by the resolution of the rendered images, thus may
lead to missing details such as deformation for tiny garment pieces and small wrinkles. In addition, we only consider pose-dependent deformations, while omitting complex dynamics such as interactions between multiple layered garments.

\noindent
\textbf{Conclusion. }In this paper, we propose a novel skinning-free pipeline to generate high-fidelity 3D garment deformation via image transfer. We decompose garment deformation into decoupled frequency modalities represented by vertex positions and normals, and further project both modalities into the image space, which allows us to leverage powerful vision models to produce wrinkles of superior scalability and perceptual quality. Thanks to these designs, our method effectively produces finer wrinkle details over previously dominant skinning-based baselines.

\noindent
\textbf{Acknowledgement.} This research is funded in part by an ARC (Australian Research Council) Discovery Grant of DP220100800. Prof. Hongdong Li holds concurrent appointments as a Full Professor with the ANU and as an Amazon Scholar (part-time). This paper describes works performed at ANU and is not associated with Amazon.

{
    \small
    \bibliographystyle{ieeenat_fullname}
    \bibliography{main}
}

\end{document}